\documentclass{article}
\usepackage[utf8]{inputenc}

\usepackage{natbib}
\usepackage{graphicx}
\usepackage[normalem]{ulem}
\usepackage{url}
\usepackage{xcolor}

\usepackage{amsfonts}

\begin{document}

\title{Generative Art Using Neural Visual Grammars and Dual Encoders}
\author{Chrisantha Fernando, S. M. Ali Eslami, \\Jean-Baptiste Alayrac, Piotr Mirowski, \\Dylan Banarse, Simon Osindero\vspace{0.1cm}\\\small{DeepMind}}

\date{\small{March 2021}}

\maketitle

\begin{abstract}
\noindent
Whilst there are perhaps only a few scientific methods, there seem to be almost as many artistic methods as there are artists. Artistic processes appear to inhabit the highest order of open-endedness. To begin to understand some of the processes of art making it is helpful to try to automate them even partially. In this paper, a novel algorithm for producing generative art is described which allows a user to input a text string, and which in a creative response to this string, outputs an image which interprets that string. It does so by evolving images using a hierarchical neural Lindenmeyer system, and evaluating these images along the way using an image text dual encoder trained on billions of images and their associated text from the internet. In doing so we have access to and control over an instance of an artistic process, allowing analysis of which aspects of the artistic process become the task of the algorithm, and  which elements remain the responsibility of the artist. 
\end{abstract}

\section{Introduction}

Generative art has been defined as art partly or wholly produced by an autonomous system, but this is not very helpful because Hieronymus Bosch is an autonomous system. It is perhaps better to say that generative art can be defined as art that involves in its production some specifiable formal system, i.e. an algorithm (note that according to this definition, it need not necessarily involve a digital computer; uncooked spaghetti, a recipe, and an obedient human would suffice). Also it is not necessary, and indeed, currently, not possible that all aspects of production of the artwork be formal. Notably, in the majority of cases, there is a  human artist who plays the role of the critical evaluative agent, the computer being used to implement only the ``technically skilled'' aspects of the production process.\\

Early examples of generative artists include Manfried Mohr, Vera Molnar, Harold Cohen; more broadly, perhaps, the conceptual artists such as Sol LeWitt; and even artists such as Gehard Richter in certain pieces of work whose production could be automated and outsourced to any competent human crafts person \citep{du2019creativity}. The formal process can play multiple roles in the production of the artwork. For example it may play a role in the search for the artwork itself, as Vera Molnar has made clear in her writings about her own process \citep{10.2307/1573236}. Or it may play a more limited role in the specification of the final artwork as in much of Sol LeWitt's work \citep{lewitt1967paragraphs}.

\begin{figure}[h!]
\centering
\includegraphics[scale=0.7]{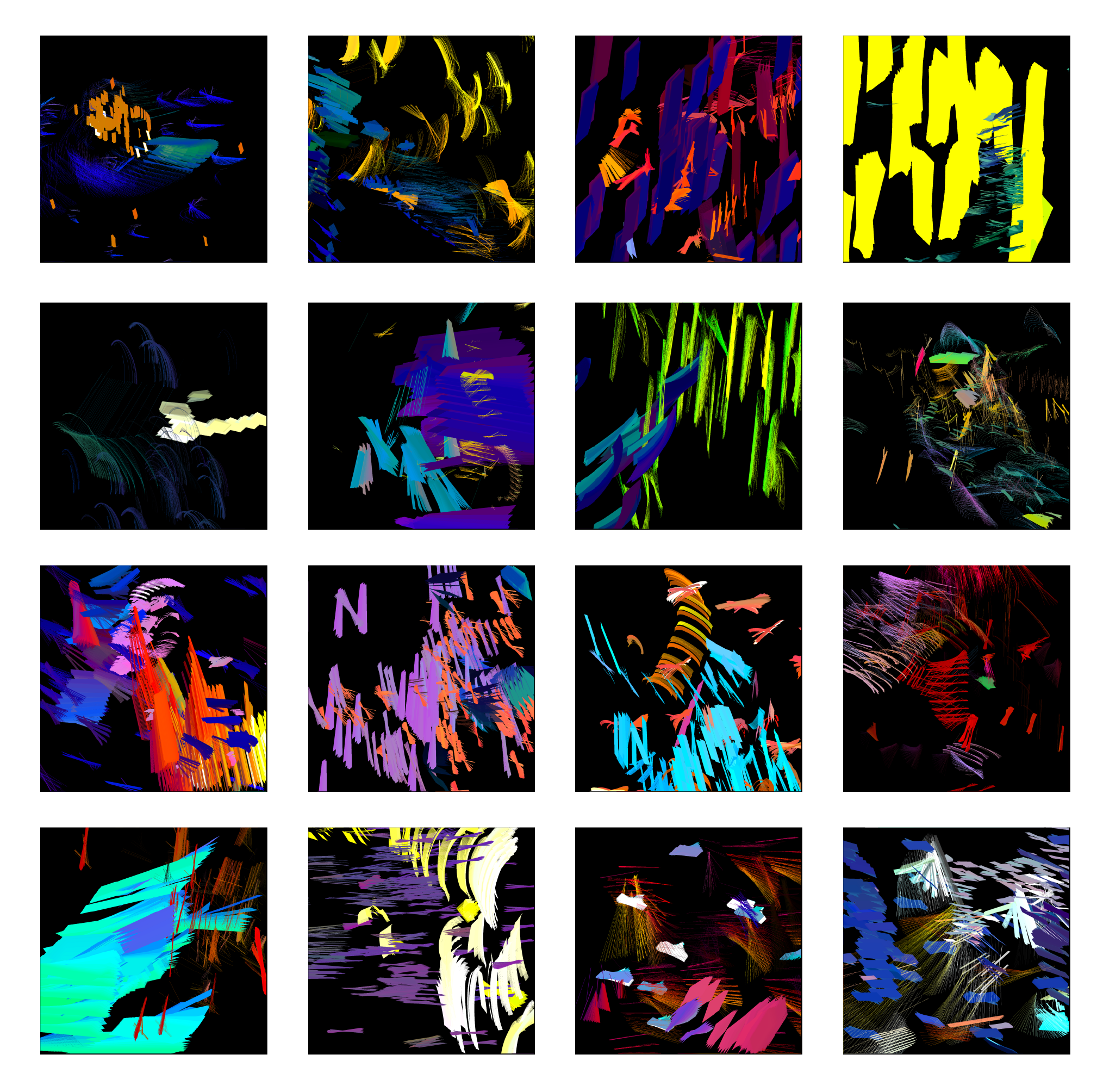}
\caption{Images produced by the final hierarchical visual grammar system for the text prompt ``Jungle in the Tiger''.}
\label{fig:0}
\end{figure}

However, the above is insufficient as a definition because it would leave the logical possibility open of saying that any art is generative art if it can be copied and perfectly imitated, e.g.,\ by a high definition 3D scanner and printer, therefore meaning that a Rembrandt painting is generative art if it can be forged using such a trivial formal process. A mildly refined definition states that an artwork is generative to the extent that it is possible to make a copy of the artwork whilst still losing information (at the pixel level) about the artwork itself, i.e.,\  if we can identify a procedure for it's construction which is more compressed than the final product (perhaps \emph{emergent-art} is a better term \citep{mcgregor2005levels}). But this may not be entirely sufficient either, because we may require that beyond the ability to reproduce an existing artwork algorithmically, whilst losing as much information as possible, we also require a piece of generative art to be made by a known formal process that can also reliably produce other equally good artworks of this same type when sampled or reinitialized. But this even stricter definition may well leave no examples of existing algorithms that produce consistently satisfactory generative art, because in most cases the result will have been evaluated as unsatisfactory by the artist. \\

The point we would like to stress is that the production of an artwork is a complex process, some parts of which have been and can be made algorithmic. Some companies such as Adobe do this very well with their `Creative' products for photo or video editing, illustration, character animation, etc. Some parts of the artistic process though, have not yet been made formal or algorithmic, such as for example the motivation to produce art by the artist, or the process by which art is judged as worthy by the artist, or even by the art critic or the collector. Crudely speaking in Aristotelian terms, the \emph{material cause} can be dealt with by paint or digital screens etc., the \emph{efficient cause} by the crafts person (or Adobe software), but the \emph{formal causes} remain a challenge to automate, and the \emph{final causes} remain an almost complete mystery. It is not possible to fully automate the artist, in the absence of some kind of motive to express something. What would give a system a credible way to possess this kind of motive is the critical question. What process would allow a system to engage in the pragmatics of making art in various social contexts? What would make it freely able to choose to make art and why would it do so? These are not questions we address here, but which some of us believe (perhaps unsurprisingly as we work for an AI company, in contrast to Aaron Hertzmann a scientist and artist at Adobe Research  \citep{hertzmann2020computers}) that these are challenges which can eventually be addressed algorithmically. In any case, a generative artwork is an interesting example of a liminal artifact at the interface of the formal and non-formal worlds, and that is why it is an interesting object for functionalists interested in the limits of what artificial general intelligence can be. But now back to basics.\\

In the field of machine learning, the (far from cottage) industry which is Generative Adversarial Networks (GANs) \citep{goodfellow2014generative} adds a critical evaluative element to the art making process, by training a neural network discriminator as a critic which must evaluate whether a computer generated image comes from the same class as the images from a human curated dataset (or not). At the same time a generator neural network is trained to fool this critic into thinking that the image it produces is indistinguishable from an image from this image dataset chosen by the human curator \citep{goodfellow2014generative}. To produce art with GANs then, the human is displaced one step further in the creative process, than if they had to construct the image themselves without any formal aid. Their creative act is once removed in that they become the curator of the dataset used by the discriminator, rather than making minute by minute decisions about which mark is good and which mark is bad (or which creative decision more broadly is good or bad). For the discriminator, these micro-evaluations that are made in producing an artwork are formalized into the well defined question: is this image distinguishable from the dataset the curator chose for me or not? Therefore, decisions about this dataset massively influence the images produced, and it is at this level that the generative artist will mainly operate. The skill of the generative artist here is at the level of choosing the algorithmic and architectural constraints of the GAN, and choosing the dataset. An interesting variant to the GAN approach is the Creative Adversarial Network (CAN) which adds a cost for making images which are too easily classifiable as an existing artistic genre \citep{elgammal2017can}. However, this process is liable to produce works that resemble half remembered Chimerias of two or more movements e.g. half Impressionism half Op-art, rather than producing out of distribution novelty of any real depth of understanding, which we believe is possible only within an evolutionary or reinforcement learning framework. \\

A related approach which also automates the critical evaluator is to generate images that fool a pre-trained image classifier into giving a high probability to them being images from the class used for training the classifier. Usually gradient-descent backpropogation is used to train the system end-to-end \citep{mordvintsev2015deepdream}, but evolutionary methods have also been used \citep{white2019shared}.  Another approach to automating the evaluative and critical aspects of the artist is taken to some extent by Simon Colton in ``The Painting Fool'' \citep{colton2012computational}, which e.g.\ produces portraits modified by its ``mood'' that is based on reading newspapers that morning (although its ``emotional'' system is highly limited: for example it does not dwell on making pessimistic predictions of the future given what it reads in the Daily Mail, nor does it contemplate the suffering and miserableness it most likely must later endure).\\ 

Here we utilize a new kind of automation that has recently been developed. We focus on making an algorithm that produces interesting images based on general linguistic prompts. We acknowledge that art need have nothing to do with images, but it is a start. Our approach differs in three ways from the use of GANs to produce images. First we evolve images, rather than using backpropogation to generate them, as in a recent paper which backpropogated through a differentiable Scalable Vector Graphics (SVG) system \citep{li2020differentiable}. Using evolutionary-search, results in a very different aesthetic output, and also allows far greater human control over the aesthetics of the output. Second, we do not evolve the images directly, but instead evolve a visual grammar for generating images. This results in interestingly structured images, e.g.\ with repetition of forms with variation. Thirdly, we use a pre-trained multimodal critic that has been trained on a vast array of both images and captions from the internet. This novel system relies on very large datasets with visual representations matched to words defined culturally on the internet. The extent of this system's ``understanding'' is important. It is not just a bag of words representation, but understands the grammar of sentences to some extent. Word order is important. Such networks have been shown to have some semantic compositionality, understanding the visual meaning of short image captions. Therefore they can be used to select for novel combinations of images that did not previously exist. An entirely gradient-based version of this process has recently been published by OpenAI \footnote{An approximated implementation by Ryan Murdoch, employing CLIP \citep{radford2021learning} critics and Big GAN \citep{brock2019large} generators, has been made available at \url{https://github.com/lucidrains/big-sleep}} with some remarkable results \citep{ramesh2021zeroshot}. We replace the generative process used there with an evolutionary process over which we can exert artistic control. Human artistic evaluation is here displaced to the determination of which generative processes are interesting, and then to exploring which captions are able to generate interesting images. In some cases the images produced are surprising and elegant in their simplicity, in others they are banal and a complete mess. Once a generative procedure has been decided upon, the role of the human user is to choose the textual prompt carefully so as to produce the kinds of images that are desired. The human user can then do with these images as they will. \\

Critically, many of the decisions about how to visually represent a word or sentence are not made by the human at all, but are made by the generative and evaluative machine. This means that various cultural references may incorporate themselves sometimes rather cryptically into the artwork without the knowledge of the artist. Historically, it has almost always been the case that the representational decisions were made by the artist, but this has now been outsourced to the computer to some extent. This is quite a thing to outsource, and it's outsourcing is largely a novelty in the artistic process, the consequences of which are not fully appreciated yet. With it come certain justifiable fears such as the fear that the critic may make representational decisions which had they been made by the artist may have disgusted and appalled them. After all, our automated critic provides no explanation of why they chose to reward a certain representational decision rather than another. \\

\begin{figure}[h!]
\centering
\includegraphics[scale=0.55]{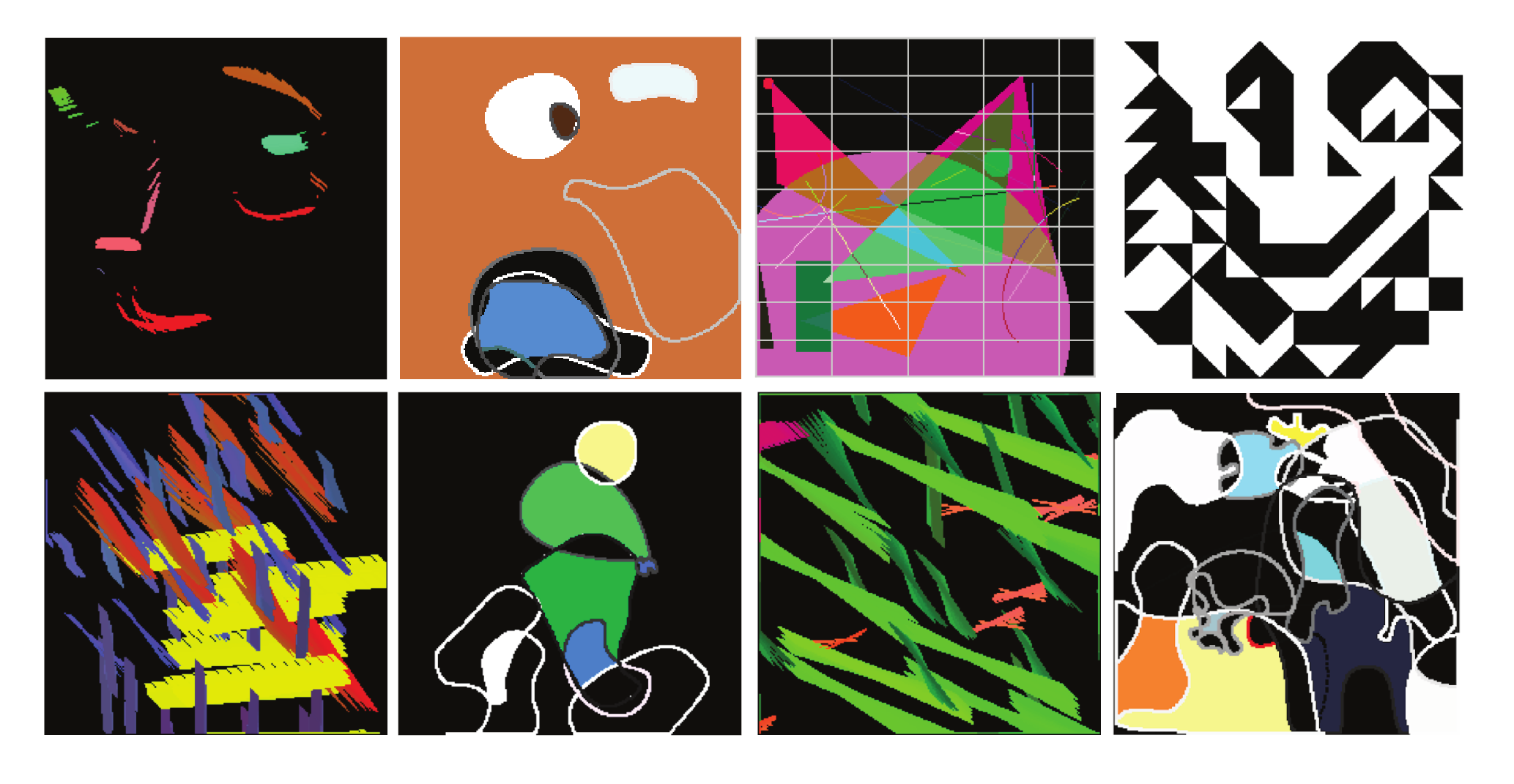}
\caption{Some examples of evolved images for the sentences (from top-left to bottom-right respectively: ``a face'', ``scream'', ``a cat'', ``a smiley face'', ``a house on fire'', ``a person walking'', ``a tiger in the jungle'', ``a cave painting''.}
\label{fig:1}
\end{figure}

What is this critic? Let us consider the multimodal transformer in more detail. A recent advance in deep learning allows the creation of novel images from titles, captions or descriptions of the image. For example if one types ``An armchair in the shape of an avocado" the system is able to generate pictures of that scene \citep{ramesh2021zeroshot}. The system ``understands'' visual concepts such as colour, texture, shape, object relations, perspective, and style, among others, to the extent that it is able to generate plausible images for combinations of these terms in novel input sentences that have never been seen before. The images produced are remarkably convincing. The system is trained by supervised learning based on billions of images and their (possible) captions from the internet (these systems are known as multimodal neural networks because they are trained on two modalities) using a ``transformer'' architecture which implements associative learning in a new and efficient way \citep{dosovitskiy2020image}. Related neural networks, dubbed image text dual encoders, are able to give a score to an image based on how closely it resembles a sentence \citep{radford2021learning} and it is this kind of network we use here as our critic. We use this critic to evaluate the {\em fitness} of an image; the higher the fitness the closer the resemblance. We then evolve images that we generate using stochastic search over abstract image primitives, rather than using gradient backpropogation to produce a greedily optimized image which would maximize the probability of a sentence, as has been done elsewhere \citep{45507}. The final evolved image is a result of both interaction between what images can be easily produced by the generative process (style) and which images are scored highly by the evaluative process (representational content). The existence of the critic means it is no longer necessary to use human evaluation to evolve images as with e.g.\ Picbreeder \citep{secretan2011picbreeder}, and a much more general set of images than just ImageNet object classes can be evolved \citep{white2019shared}. Figure \ref{fig:1} shows a sample of images produced by a variety of generative algorithms we designed at the early stages of our project. The details of the generative processes and the evolutionary algorithm used to optimize the images are described in Methods.\\

\begin{figure}[h!]
\centering
\includegraphics[scale=0.55]{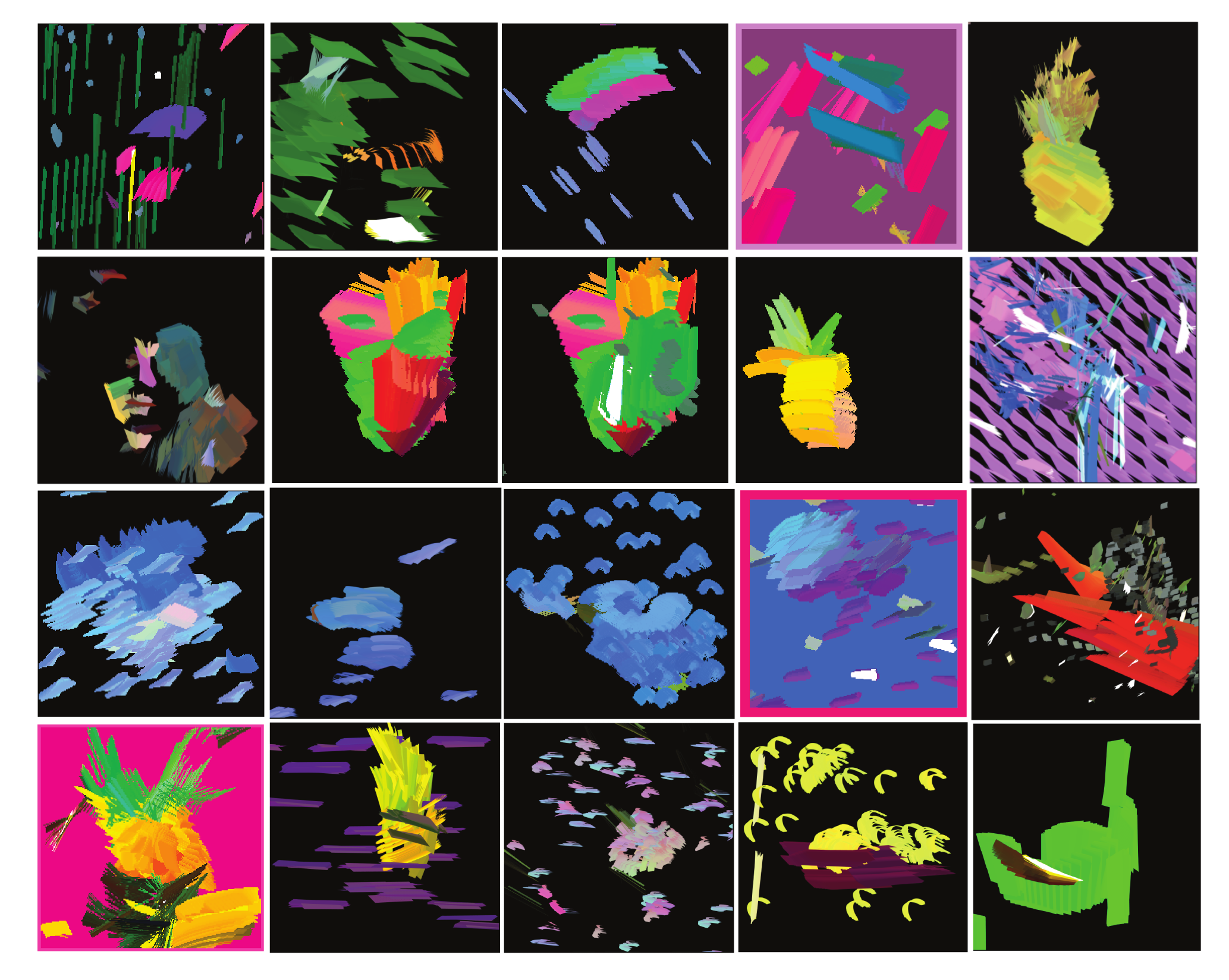}
\caption{Examples of images evolved using the final hierarchical visual grammar generator. {\bf Top row} (from left to right): ``Spring rain'', ``Tiger in the jungle'', ``A rainbow'', ``Red bermudas'', ``A pineapple''. {\bf Second row}:``A self-portrait'', ``A Sri Lankan mask'', ``A Sri Lankan mask'' (re-evolved with marks on top of the previous image), ``A pineapple'' (with B\'{e}ziers allowed), ``David Hockney''. {\bf Third row}: ``Clouds'', ``Clouds'', ``Clouds'', ``Clouds'', ``A plane crash''. {\bf Bottom row}: ``A pineapple'' (allowing background colour evolution), ``A pineapple crash'', ``Takashi Murakami'', ``A Sri Lankan cashew nut curry'', ``An armchair in the shape of an avocado''.}
\label{fig:BestExamples}
\end{figure}

The main finding of this work is that a hierarchical image generation system (which can be thought of as a neural Lindenmayer system \citep{lindenmayer1968mathematical}) can be evolved to produce a wide variety of structured images in conjunction with an image text dual encoder critic. Figure \ref{fig:BestExamples} shows examples of the kind of final images we evolve from scratch within 100,000 binary tournaments using the visual grammar generator. In the sections that follow we describe the process, provide some examples, and discuss further implications of our work.\\

\section{Methods}

\subsection{The overall evolutionary framework} 

\begin{figure}[h!]
\centering
\includegraphics[scale=0.62]{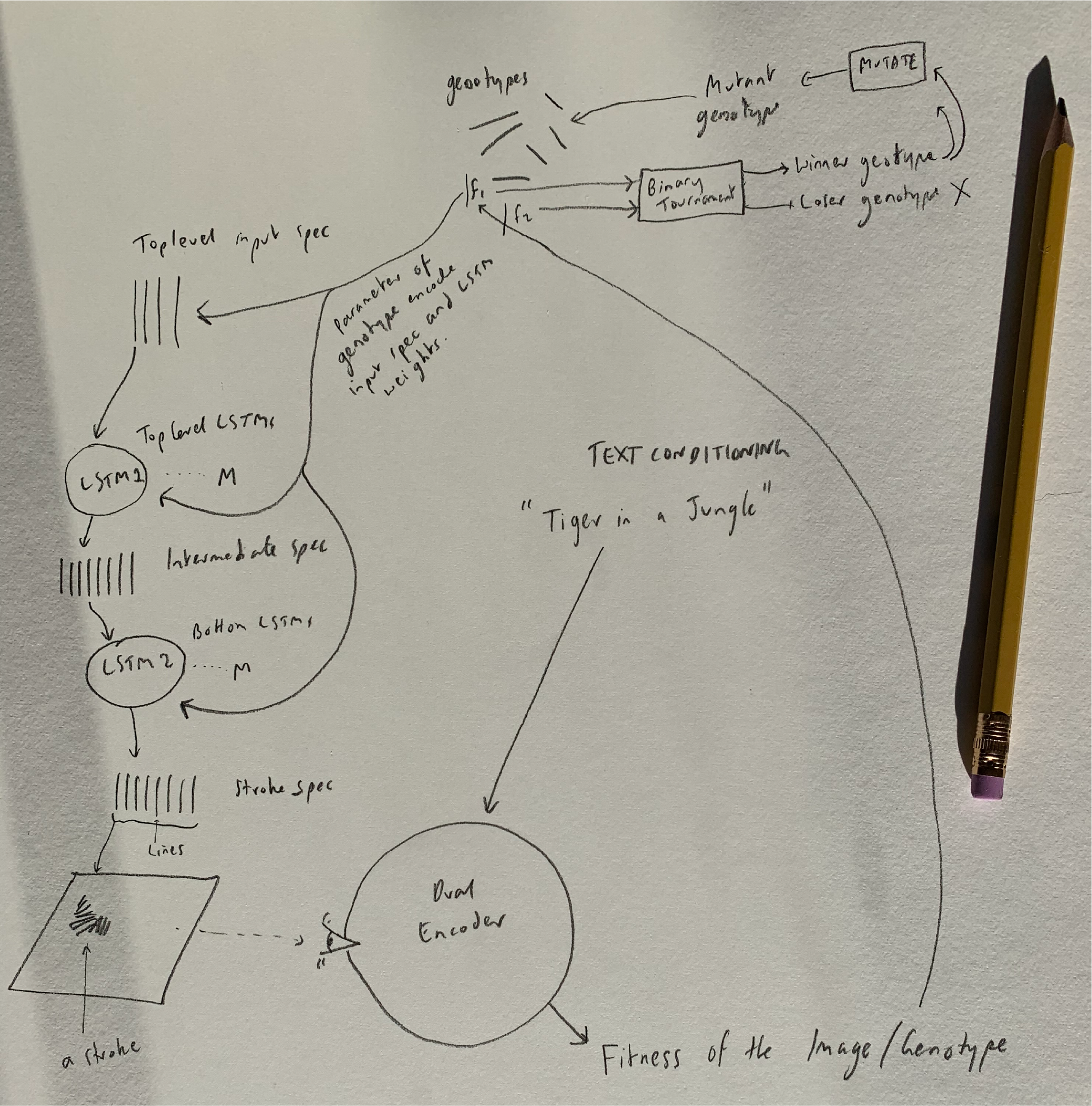}
\caption{The overall process involves first producing a random population of genotypes. Two random genotypes are evaluated. This is done by translating the genotypes into an input string and the weights of the LSTMs, and using this system to produce the image. The image is then viewed by the Duel Encoder which also recieves the text input, in this case "Tiger in the Jungle", and returns the fitness of the image. This fitness is used to determine which of the two genotypes won the competition. The winning genotype is then mutated and the mutant returned to the population overwriting the losing genotype.}
\label{OverallMethod}
\end{figure}

A parallel asynchronous binary tournament selection algorithm is used to evolve the images \citep{harvey2009microbial}. Evolutionary experiments start with a population of randomly initialised {\em genotypes} and performs 100,000 {\em binary tournaments}. Each genotype encodes all the information required to construct an image, namely an input string to a recurrent neural network and the parameters of a recurrent neural network. A binary tournament consists of the evaluation of two genotypes followed by replacement of the genotype with lower fitness by a mutated copy of the higher fitness genotype. An evaluation of a genotype involves first the construction of the image from the genotype followed by scoring the image (assigning fitness) using a dual encoder conditioned on an short input text. Figure \ref{OverallMethod} shows a representation of the overall method.

\subsection{Generative and mutational mechanisms} 

We experimented with a wide variety of generative algorithms to produce the images, see Figure \ref{fig:1}. They differ in how rapidly they can be evolved to high fitness. A good representation is one which can be rapidly evolved by the critic to high fitness. One very effective method we found involved the specification of smooth closed B\'{e}zier shapes followed by specification of flood fills at particular points in the image, see Figure \ref{fig:3_2} which shows this method to evolve an image that CF's daughter requested be included. Less effective methods include specification of geometric primitives and their colours, or the encoding of Tangram-like shapes on a grid. The least effective method is to specify individual pixels because doing so allows the evolution of extremely adversarial examples \citep{nguyen2015deep} which do not resemble the image at all to a human observer, but which score highly according to the transformer. This was limited by applying random projective transformations on each image evaluation, a procedure we adopt for all generative procedures subsequently. We did not experiment with Compositional Pattern Producing Networks (CPPNs) which have also been used in this context \citep{secretan2011picbreeder} because their aesthetics are canalized. \\

\begin{figure}[h!]
\centering
\includegraphics[scale=0.55]{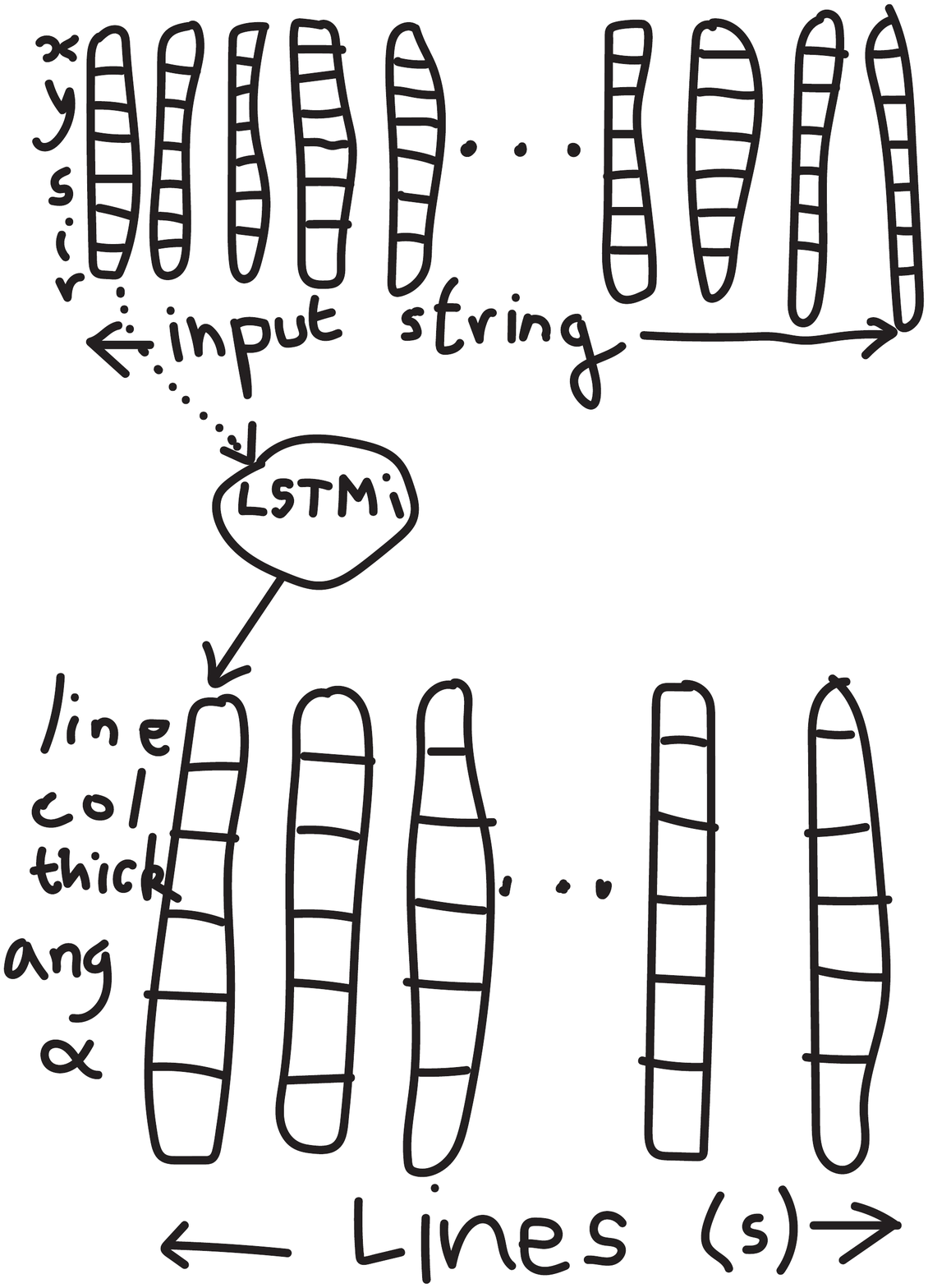}
\caption{Architecture of the single layered image generator.}
\label{fig:Arch}
\end{figure}

The core neural image generator architecture is shown in Figure \ref{fig:Arch}. It starts with the evolved input string at the top which is a list of L column vectors of length 10 each. L is uniformly distributed from 10 to 100 initially. Each vector is input into one of $M$ Long Short-Term Memory networks (LSTM, a type of recurrent neural networks \citep{hochreiter1997long}) to produce a stroke. A stroke consists of a sequence of $s$ drawn lines. Some positions in the input vector (at the top of the figure) have a specific meaning. Position 0 and 1 determines the $(x,y)$ origin of the stroke on a Cartesian grid description of the image. Position 2 determines the length of the stroke $s$, which corresponds to the number of steps that the LSTM will be allowed to iterate with that column vector as input. The 3rd position specifies the index $i$ of the LSTM that this vector will be input into. The 4th position $r$ specifies whether this stroke will be opaque or transparent. The remaining positions are evolvable inputs into the LSTM but have no pre-specified meaning. Each of the M LSTM modules outputs a vector of length 16. The positions in this vector have specific meanings which are used to describe a single line of the stroke. A stroke consists of s lines. How does the output vector specify a line in the stroke? The meanings of the output vector are as follows: Position 0,1,2,3 specify the start and finish x,y displacement of the line from the origin specified in the x,y positions in the input vector at the top. Position 4,5,6 specify the RGB value of the line. Position 7 specifies the line thickness which is how often this line will be repeated in a range from 0 to 35. Position 8 and 9 specify the inputs to a sine and cosine function respectively which determine the angle at which the line will be repeated, i.e.\ this determines the angle of an italic looking line. Position 10 determines the extent of colour mixing with existing colours if the stroke is specified as transparent by the input vector at the top. Note that the system is entirely ballistic, not referring to the image at all in its production, and so can be batched for efficiency of generation. 

\subsection{Deep Neural Image Grammar} 

\begin{figure}[h!]
\centering
\includegraphics[scale=0.4]{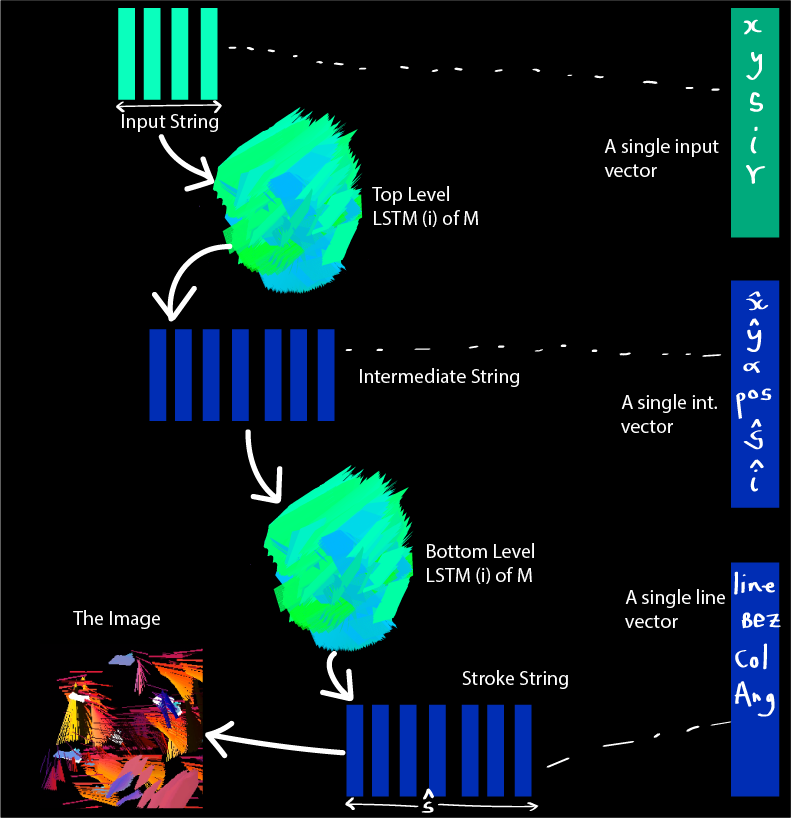}
\caption{Architecture of the deep grammatical image generator.}
\label{fig:Arch2}
\end{figure}

The hierarchical architecture which we found considerably extends the capabilities of the core neural generator is shown in Figure \ref{fig:Arch2}. It implements to some extent the kind of thing that Zhu and Mumford were talking about in their paper ``A stochastic grammar of images'' \citep{zhu2007stochastic}. However, there is a critical insight which is the ability to define the `what' of the entity produced at a level independently from the `where' of the entity at that level. This time, instead of the input string specifying the strokes directly, it is input into a top level LSTM which specifies an intermediate input string for each stroke. This intermediate input string acts very much as the original input string did. It is input into a lower level LSTM which then outputs the stroke description as before. The vectors in the original input string at the top continue to have specific meanings. Position 0 and 1 now encode the potential x,y origin position of the ``object'' to be drawn by this vector. Whether this top level encoded position is directly used or not is determined by the output of the LSTM (as we will see later). Position 2 of the top-level input vector encodes the length (s) of the intermediate input string to be produced by the top level vector, i.e.\ how many steps the top level LSTM will be run for with this vector as input. Position 3 of the top-level vector encodes which LSTM (i) will be used to interpret this vector. The remaining positions in the vector have no explicit meaning but are input to the top-level LSTM and are evolvable. \\

The output of the top level LSTM is a set of s intermediate vectors for each of the top level vectors. Some positions in the intermediate vector have specific meanings. Positions 0 and 1 determine the x,y origin of the stroke if it is determined that the stroke position will be specified at this level rather that at the top object level. The 2nd position of the intermediate vector determines whether the stroke it encodes will be opaque or transparent. The 3rd position determines whether the top level specified position or the positions specified at the intermediate level will be used to determine the origin of the stroke. The 4th position determines the number of lines to be produced in the stroke. The 5th position determines which of the lower level LSTMs will produce the stroke. \\

The encoding of lines in the stroke at the lowest level is very much like before, except this time we also encode the positions and control points and curvature of a B\'{e}zier curve, and an extra position which specifies whether a B\'{e}zier or a line should be drawn. \\

In this way, a single vector at the top level specifies an ``object'', which is defined by a list of s1 intermediate vectors which specify a set of s2 strokes that make up that object, with the additional ability for top or intermediate level vectors to independently specify where that object should appear. We find that each LSTM encodes a specific kind of stroke, and that the top level string is capable of determining where each LSTM should make it's object. All the parameters of the top level strings and the LSTMs are evolved using mixed Gaussian and Cauchy mutations. Mutations include growth and shrinkage of the top-level string, so that genotypes can be of variable length, adding and removing lines, strokes, or objects. Variation thus exists at multiple levels of granularity. Certain kinds of fitness augmentation are experimented with. For example, a style cost is sometimes used which rewards the use of either fewer or greater numbers of stroke, but we find this not to be critical. \\

\subsection{The Image Text Dual Encoder Critic}

To play the role of the critique, we need a scoring mechanism that can assess how well an input text query semantically matches an input image.
To that end, we choose a dual encoder approach~\cite{frome13devise} that has recently demonstrated great success when trained on a large collection of web data~\cite{jia2021scaling,radford2021learning}.

The Dual encoder model consists of two encoders that respectively operates on text and images.
More precisely, the image encoder $f$ takes an image $x$ and embeds it in a vector $f(x)\in\mathbb{R}^d$.
Similarly, given an input text query $y$, the text encoder $g$ extracts a corresponding vector representation $g(y)\in\mathbb{R}^d$.
The matching score between the image $x$ and the text $y$ is obtained by taking the cosine similarity of the respective embeddings: $\frac{f(x)^\top g(y)}{\Vert f(x) \Vert_2 \Vert g(y) \Vert_2}$, where $\Vert . \Vert_2$ is the $L^2$ norm.
Given training pairs of matching image and text, the parameters controlling $f$ and $g$ are optimized so that the similarity of these pairs is high while the similarity of generated negative pairs is low.
We experimented with a variety of pre-trained dual encoders, e.g.\ CLIP \cite{radford2021learning}, but did not conduct a systematic investigation into their differences. 

If not specified otherwise, we used a dual encoder trained on the ALIGN (A Large ImaGe and Noisy-text) dataset~\citep{jia2021scaling}.
The vision encoder $g$ is a NF-Net-F0 model~\cite{brock2021high} that takes as input RGB images at resolution $224\times 224$, extract the pooled representation from the last convolutional layers and adds a linear projection to the embedding space of dimension $d=512$.
The text encoder is a 80M-parameter causal Transformer~\cite{vaswani2017attention} with 6 layers, hidden size of 1024, $12$ attention heads with keys/values dimension of $64$, intermediate dimension of $4096$ for the feed forward layers, and using layer norm before the attention and feed forward layers as suggested in~\cite{radford2019language}.
The representation is obtained by taking the activation vector of the last layer of the transformer at the \texttt{[EOS]} token position, followed by a linear projection to the joint embedding space of dimension $d=512$.

This model was trained from scratch on the same data and with the same contrastive loss as in the ALIGN paper~\citep{jia2021scaling} for approximately 2 epochs with a total batch size of 16,384. Note that the ALIGN dataset contains 1.8B image-text pairs of captioned images, with minimal filtering of captions, and that the causal (i.e., forward in time) text encoding preserves, among others, the order of the words as well as distinctions between uppercase and lowercase, meaning that text strings ``Jungle in the Tiger`` and "a tiger in the jungle'' are expected to have different language embeddings, and thus to be associated to different image embeddings.

\section{Results} 

\subsection{General principles in Image Evolution} 

\begin{figure}[h!]
\centering
\includegraphics[scale=0.55]{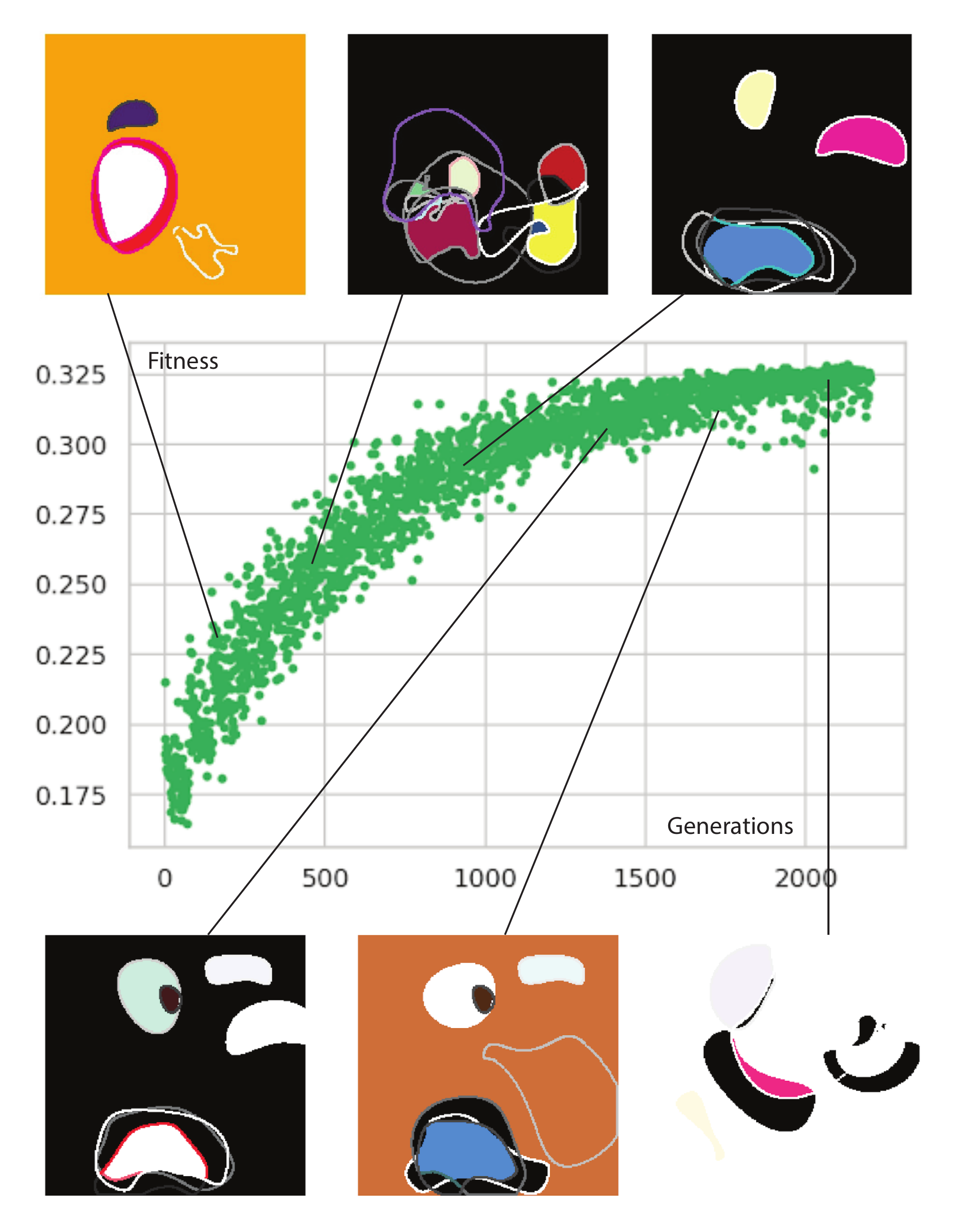}
\caption{An evolutionary run for the title ``scream''. Representational veridicality increases but eventually abstraction dominates due to perhaps a kind of pathological overfitting.}
\label{fig:2}
\end{figure}

As usual in any evolutionary run, there is rapid loss of diversity with convergence to a few instances from which more intense exploration is then carried out. This itself mirrors an artistic process of initial preparatory exploration with later exploitation. Figure \ref{fig:2} shows samples of images evolved to satisfy the title ``scream''. In many cases convergence to the final image is achieved rapidly, after which only small modifications to the image are produced. In this case the image continues to improve its score although to the human observer it begins to resemble the title ``scream'' less. This may be due to some form of pathological overfitting, and therefore after this, random projective transformations were made on each evaluation of the genotype. Perhaps the tendency to produce increasingly abstracted art is also a kind of conceptual overfitting. \\

\begin{figure}[h!]
\centering
\includegraphics[scale=0.55]{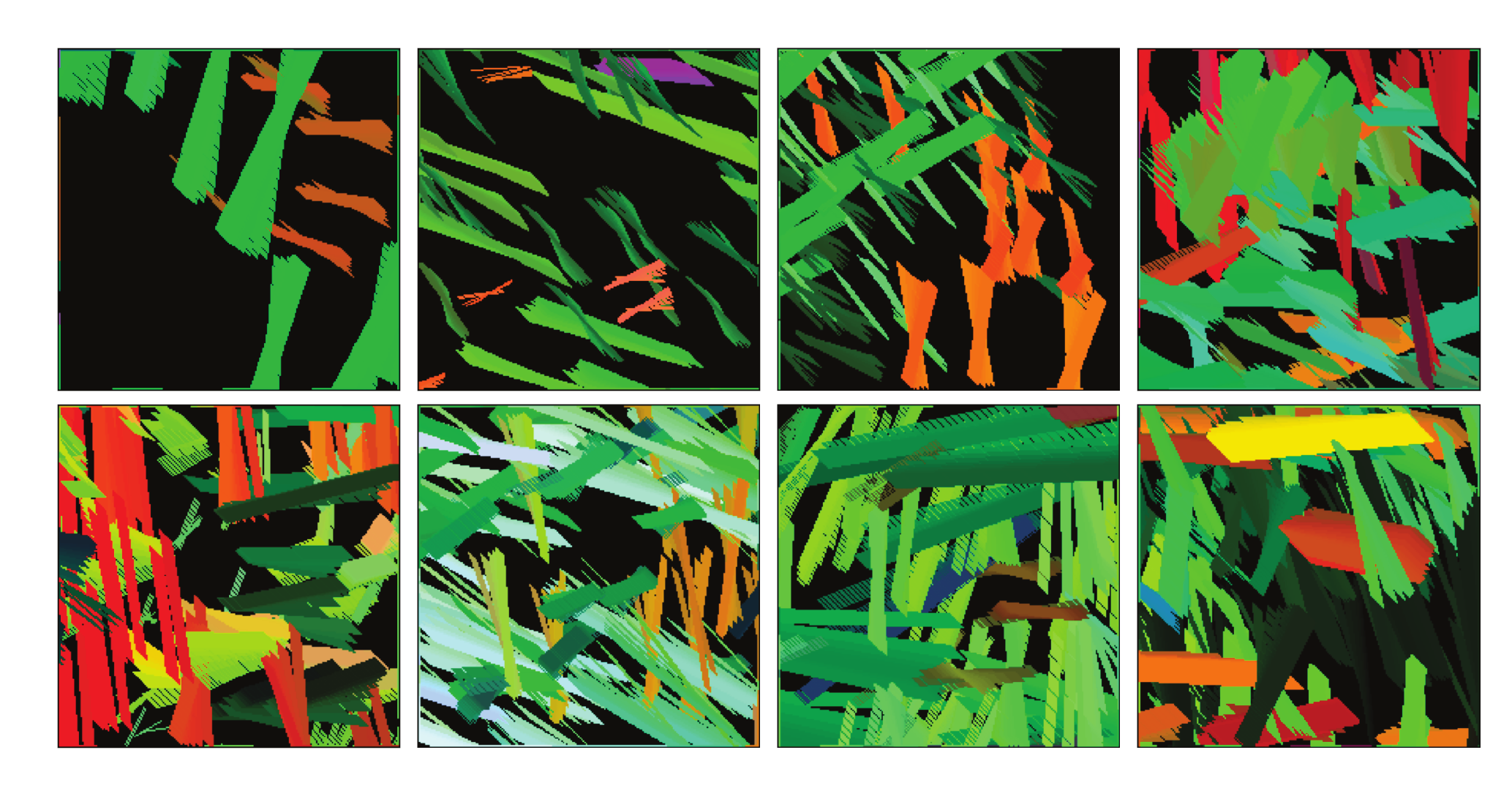}
\caption{Examples of images evolved for ``Tiger in a Jungle" on several independent runs.}
\label{fig:3}
\end{figure}

\begin{figure}[h!]
\centering
\includegraphics[scale=0.7]{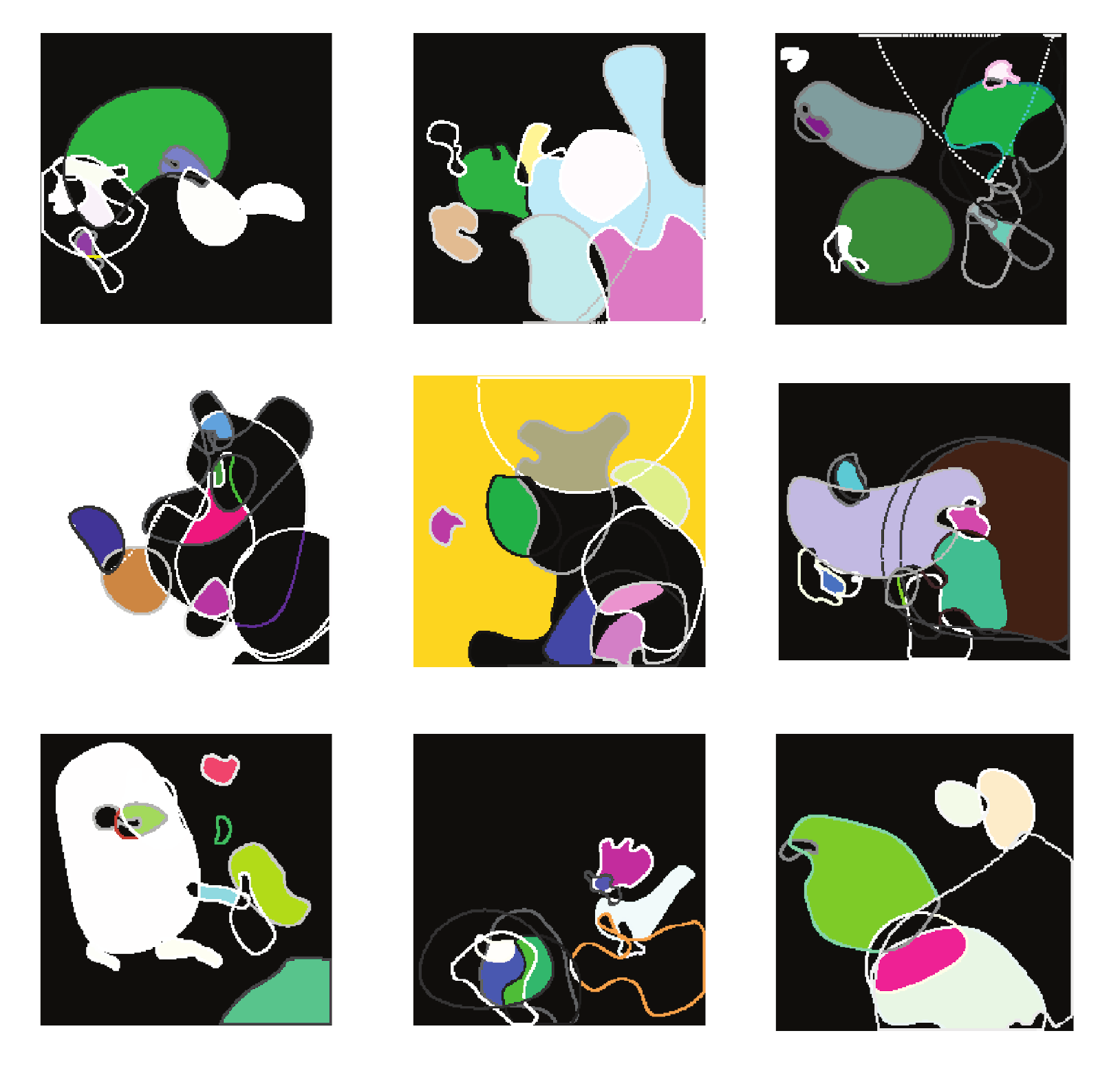}
\caption{Examples of images evolved for ``Cute fluffy dog with a parrot sitting next to it'', during a single run.}
\label{fig:3_2}
\end{figure}

What happens if the tape is run twice? Each evolutionary run is different, and even with the same text conditioning there is convergence to different visual attractors. This is wonderfully demonstrated in Figure \ref{fig:0} where all the images were produced by the same prompt ``Jungle in the Tiger''. Similarly Figure \ref{fig:3} shows that if evolution is rerun for the same title ``tiger in a jungle'', that various kinds of depiction are evolved (using the single level generator not the grammatical generator as in Figure \ref{fig:0}), which in this method of generation seems to focus on the colour of the scene and its general texture. Even during a single evolutionary run using the B\'{e}zier smooth shape encoding for the title ``cute fluffy dog with a parrot sitting next to it'' we see that several kinds of depiction are competing for prominence, focusing on representing various aspects of the title such as the parrot or the dog, see Figure \ref{fig:3_2}. The presence of multiple possible attractors for the same search string is clearly demonstrated. \\

\begin{figure}[h!]
\centering
\includegraphics[scale=0.45]{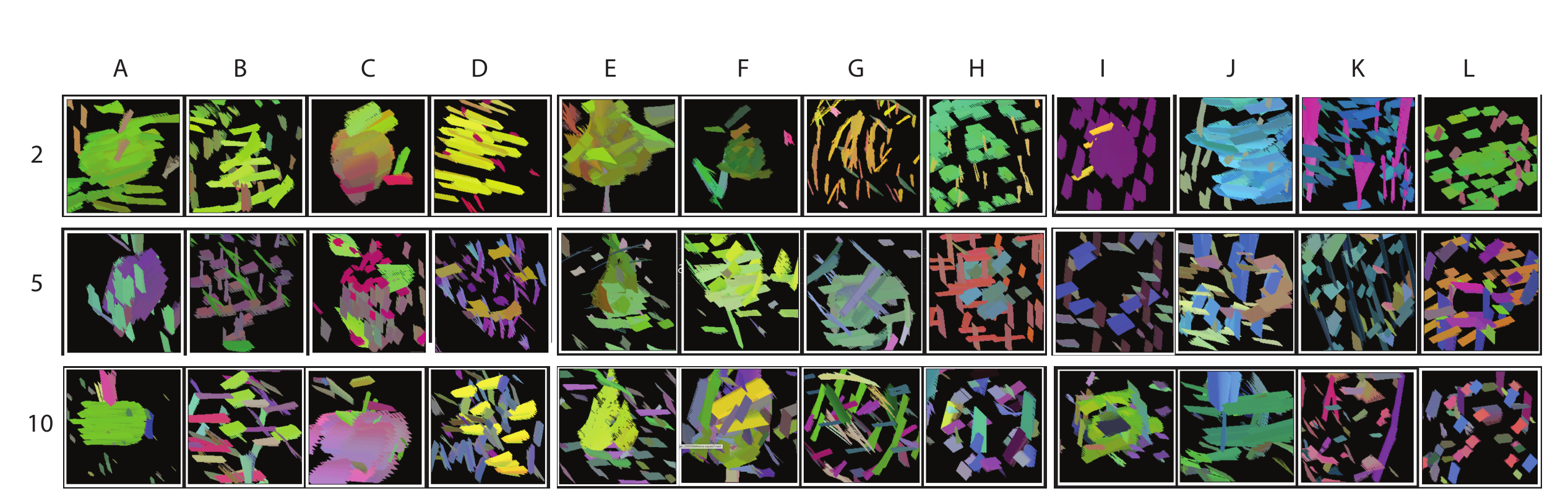}
\caption{Various titles illustrated  with modular LSTM networks of size 2, 5, and 10. A: Apple, B: a tree, C: apple, D: a banana, E: a pear, F: a lemon,  G: a circle, H: a square, I: a circle on a square, J: a glass of water, K: a glass, L: a square in a circle.}
\label{fig:3_4}
\end{figure}

Figure \ref{fig:3_4} explores the effect of slight modifications to a generative procedure. The style and feel of an image is altered by such changes. Vera Molnar describes her process of tweaking a computer program until a desired effect is achieved \citep{10.2307/1573236}. To this end, it seems the case that random generation and search for pleasing images is easier when tweaking simple programs compared to more complex ones. This is because with each added piece of complexity, there is a chance that the pleasing ordered visual Gestalt principle which was of most interest to us (and was serendipitous in its appearance) can be confused, lost, or hidden by disorder. Various attempts can be made to limit disorder such as reducing the number of independent stroke types (LSTMs) permitted. The figure experiments with a variety of LSTM numbers. Note this is still the single level system not the hierarchical system. \\

\begin{figure}[h!]
\centering
\includegraphics[scale=0.57]{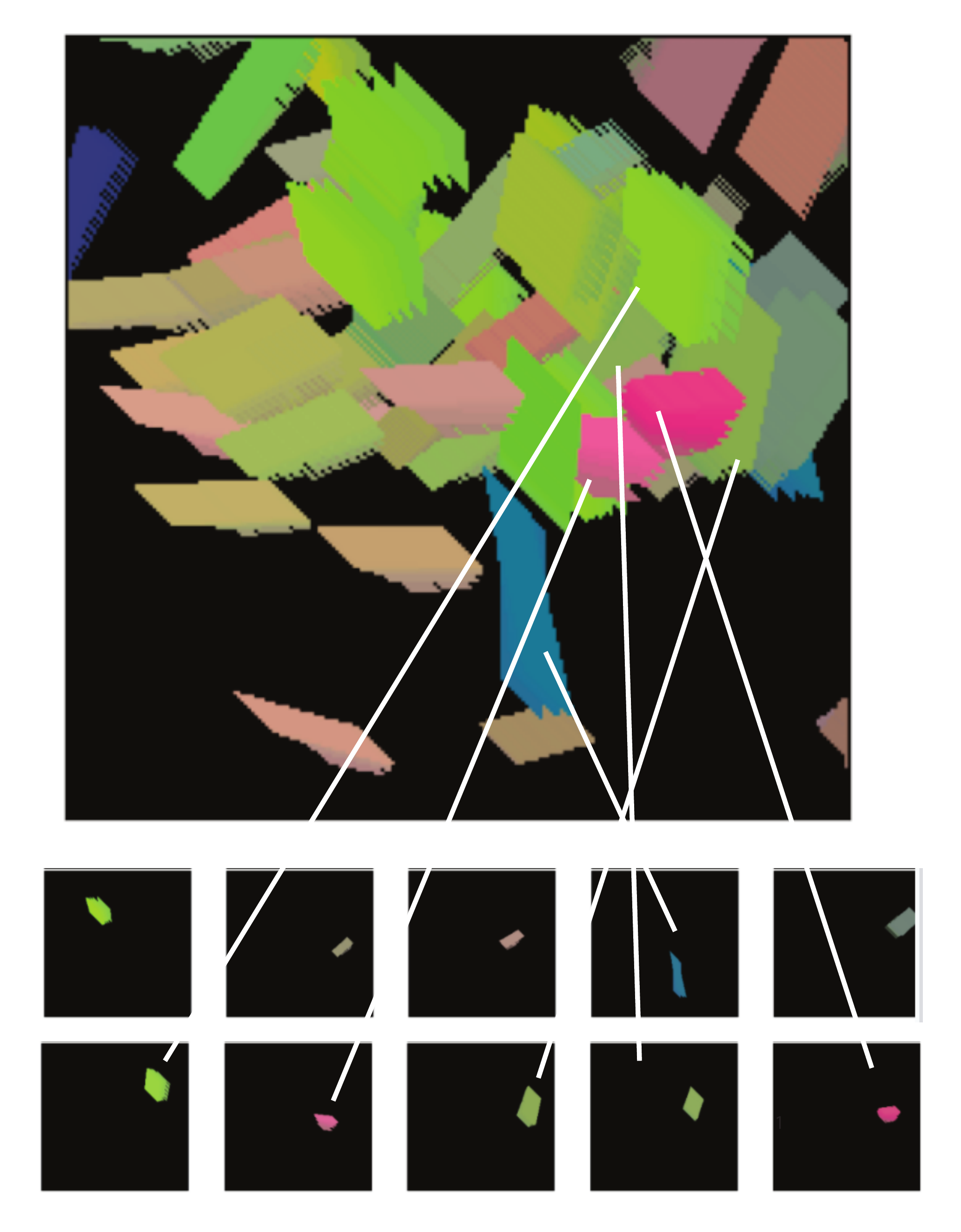}
\caption{The top ten marks which, on removal, maximally decrease the fitness of the image as an `apple' are shown in increasing importance (bottom left to top right). It seems in this case an apple tree has been drawn (which is rare, mostly an apple is drawn). The main apple in the apple tree is the most important mark, followed by some leaves, and then the trunk of the tree.}
\label{fig:4}
\end{figure}

Figure \ref{fig:4} shows how a pruning procedure can be used to determine the critical marks that contribute to the score (fitness) of an image. This procedure can be used occasionally throughout evolution to remove extraneous marks that are unnecessary for high fitness. The marks most important for recognition of the image as an apple are those that produce the stem, and the red top of the apple. Strangely there are two marks that are not obviously related to the apple such as the two purple marks on the left of the apple which are also important for the evaluator. In general we do not use this pruning procedure through evolution as it is costly and evolution alone within an asynchronous parallel system is sufficient to reduce genome sizes if useful because smaller genomes are automatically evaluated faster and replicate more even at the same `fitness' level.\\

\subsection{Evolution of Visual Grammars} 

\begin{figure}[h!]
\centering
\includegraphics[scale=0.57]{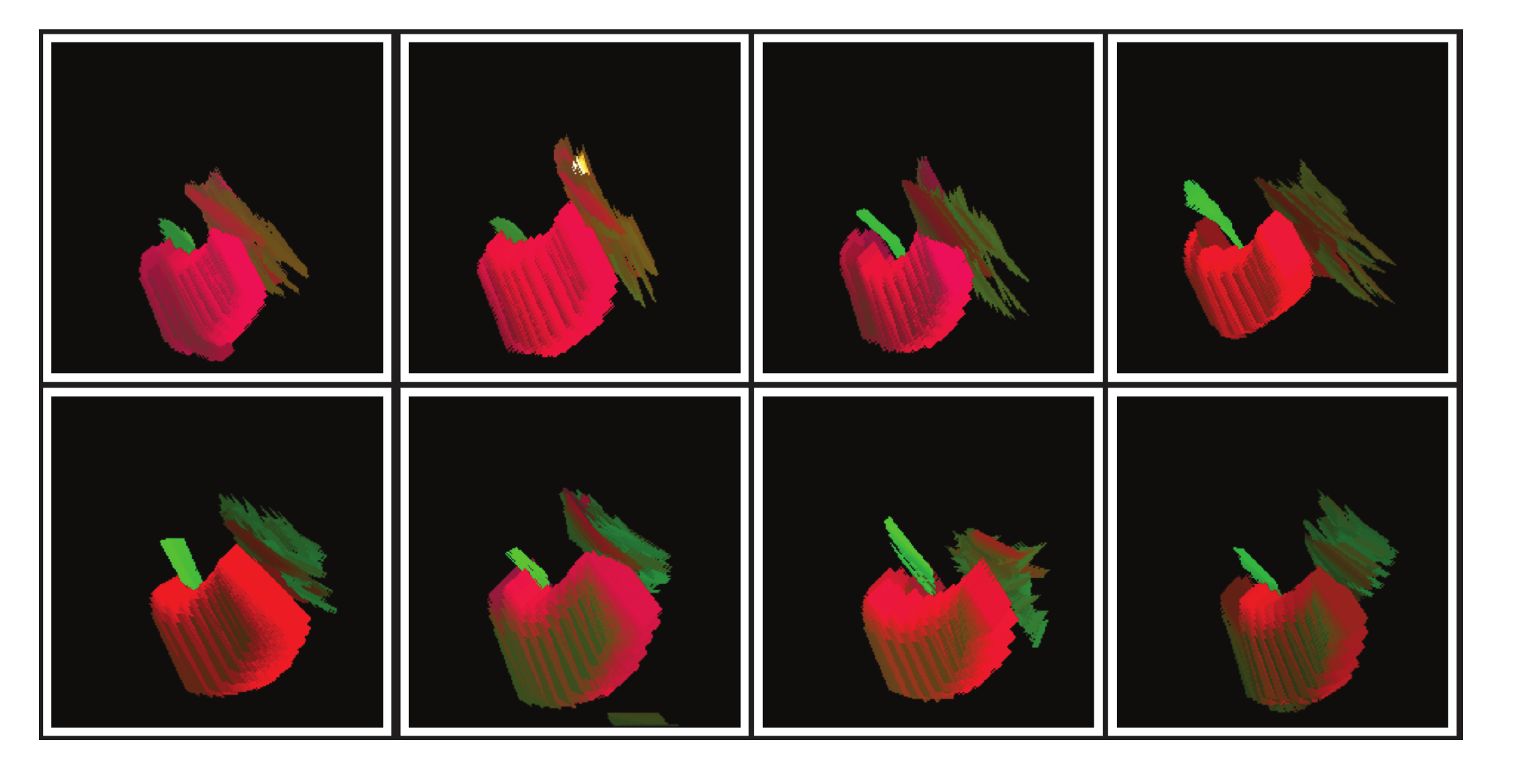}
\caption{Mutants of images for ``apple" produced by the visual grammar system}
\label{fig:5}
\end{figure}

\begin{figure}[h!]
\centering
\includegraphics[scale=0.57]{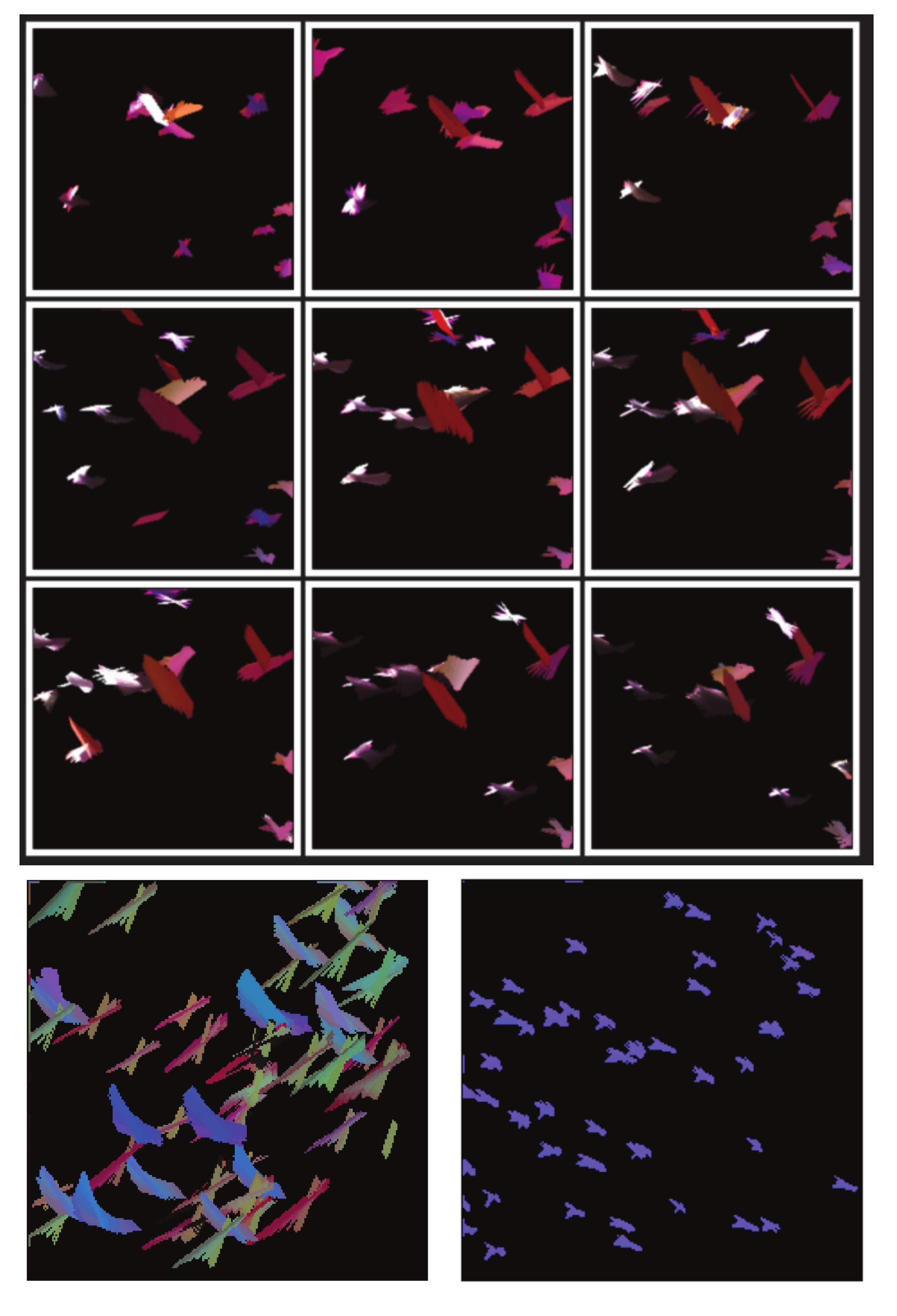}
\caption{Three independent evolutionary runs for "A flock of birds" produced by the visual grammar system.}
\label{fig:6}
\end{figure}

\begin{figure}[h!]
\centering
\includegraphics[scale=0.6]{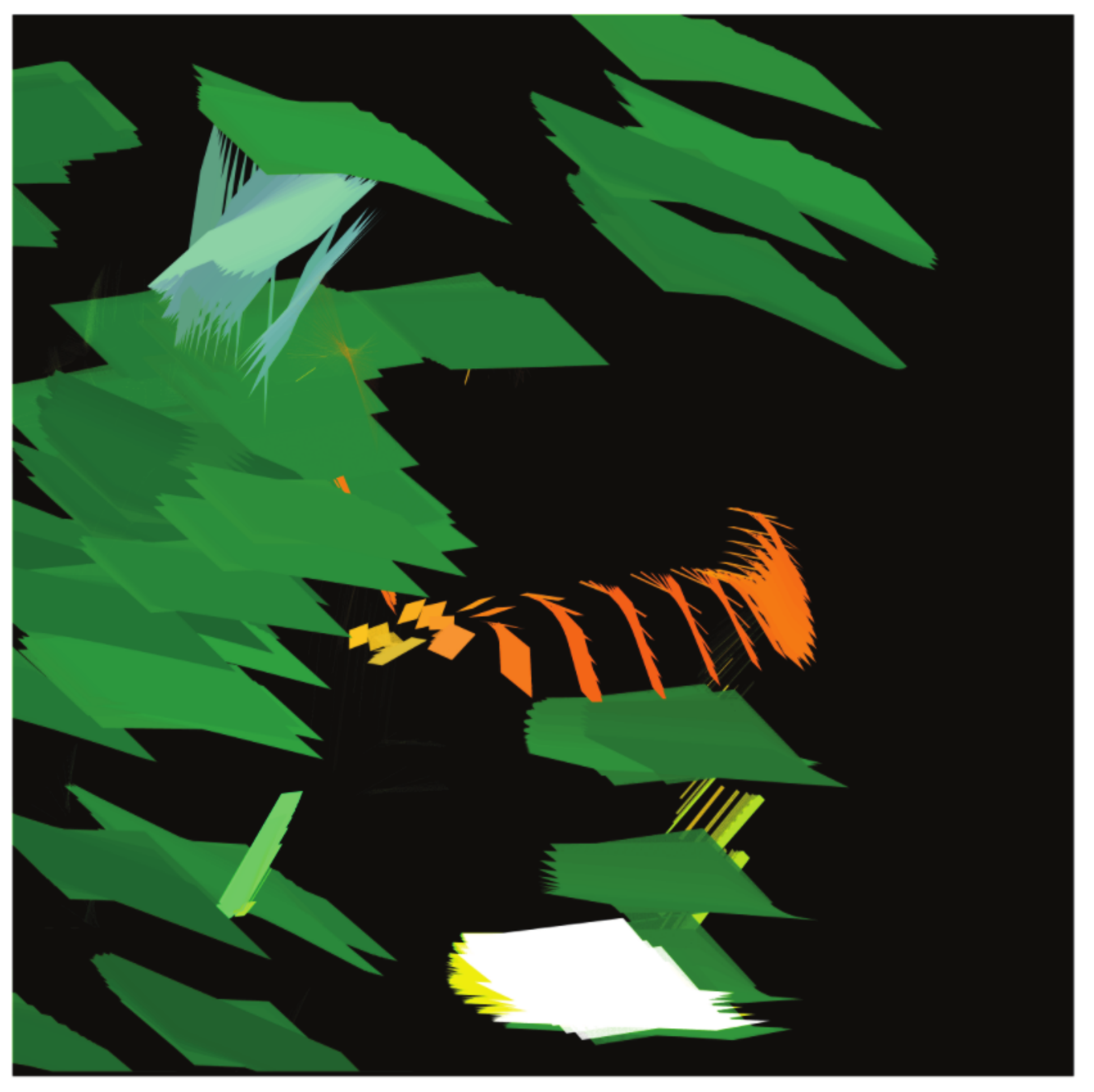}
\caption{Notice how a sequence of systematically changing strokes is used to form the back of the tiger in an elegant fashion.}
\label{fig:Tiger}
\end{figure}

Figure \ref{fig:5} shows some of the later images evolved for the title ``apple" using the hierarchical visual grammar method. They are much clearer, less random, and contain considerable structure in the variations produced by mutation. The visual grammar system is capable of repeating similar `objects' at different locations in the image, making it effective for example in producing `a flock of birds', see Figure \ref{fig:6}. Figure \ref{fig:BestExamples} shows a variety of images produced by the visual grammar system. Because we do not evolve the image directly, but the generative process, it is possible to produce high resolution versions of the image, although only the 224$\times$224 low resolution version was ever evaluated by the dual encoder. Some videos of the evolutionary trajectory of best images during a run are available here \citep{bworld}. It is interesting to observe some of the representational tricks that the generative procedure has discovered for making depictions. We particularly liked the ordered and systematic progression of stroke changes used to produce the back of the tiger in Figure \ref{fig:Tiger}. 

\section{Discussion}

We have shown how coupling a evolvable generative neural visual grammar with a multimodal transformer allows the automation of the production of simple abstracted representations on the basis of an image title alone. Some parsimonious and interesting representational methods are discovered for producing images, showing the kind of unbiased simplicity that evolutionary solutions often exploit \citep{nguyen2015innovation}. The system is an excellent colourist in particular, which was one of the major preoccupations of Harold Cohen with Aaron \citep{cohen1982make}. For example consider the colours chosen when representing a Tiger in the Jungle in Figure \ref{fig:Tiger}. Interestingly Aaron was also a hierarchical system, but was reactive to the image being produced. This is not a property of our generator, but introducing such sensitivity to what is drawn is the obvious next step. \\

The system is very different from CPPNs which have been extensively used previously to evolve images \cite{secretan2008picbreeder} in that it uses a vastly more inefficient recurrent neural network rather than generating the image in one go. The aesthetics of the images differ greatly from those produced by CPPNs. Unlike CPPNs our system seems better at rapidly evolving images that can satisfy a pre-specified target, rather than preferring to be used in neutral quality diversity mode. We believe this is because object priors are encoded into the generator with `what' and `where' like distinctions made at both levels in the hierarchy, allowing evolution to move whole objects around in the image. We sometimes also explicitly encode background and foreground priors, and border priors, which the system can choose to use if it wants, although we find this an inelegant approach.  We disturbingly noticed that often the evolution was limited to a poor local optimum by allowing the evolution of a background colour, and so we made this less likely to be discovered. It is possible that allowing a background colour to evolve rapidly selects for a single background colour, which then reduces the diversity of the genotype in other respects, making further progress hard. We also counter-intuitively found that introducing an initial random search process with selection of the initially top fit genotypes was unhelpful for evolutionary search. This is because often there are a small number of limited ways in which high initial fitness can be trivially achieved, e.g.\ using curves rather than lines, which again reduced diversity. \\

In a sense our system can be thought of as painting from memory. It is painting because of the nature of the strokes and colour mixing which alludes to a painterly tradition in art, and it is from memory because the system associates a complex conceptual amalgam of images based on historical pairings of those images with sequences of words linked to those images on the internet. GANs can be thought of as much more like photography from memory because of the more fine-grained way in which backpropogation can interrogate the discriminator. Photography from memory is a strange concept, but seems to capture concisely much of the aesthetics of GANs. \\

In the same way as GANs, our work can be seen as artistic appropriation because it invents nothing new in terms of mapping from language to image. In a sense it can only revert to the mean. To see this is true please run this thought experiment. If the multimodal transformer were retrained on images produced by the above generative process, we would not expect any new representational ability to arise, quite the opposite, a self-referential system would be likely to arise, degenerating into a few common attractors. New representational conventions such as ``the map'' or ``orthogonal perspective'' cannot be invented by our system, as indeed they cannot be invented by GANs \citep{fernando2020language}. It can only rehash existing representational conventions. To create a new representational convention it is not possible to avoid the use of a reinforcement learning system interacting with genuine behaving agents (probably humans), e.g to produce some desired behaviour through showing an image. Only in this way can truly surprising and novel visual conventions be generated. However, there is a class of novelty that our system would be able to produce in the above thought experiment, and that is stylistic novelty. New ways of depiction would arise, e.g.\ new ways of shading, new ways of concisely making a face, etc.. Further investigation of the limitations of this novelty are needed. \\

What are the implications of our system for artists and designers? The automation of any process is transformative and often disturbing. Jeff Koons' studio, where the artist was guiding a hierarchy of assistants to execute his instructions, gave a disturbing illustration of this when he replaced some of the human assistants by automated tools \citep{koons}. Possibly our application has the potential to be used as an aid to artistic practice, inventing new kinds of mark making and allowing access to representational decisions, allowing rapid simulation of possible solutions to task briefs, allowing the user to ask ``draw me a sausage in a waterfall'' with the system making its best attempt to do so. The background initial condition of the system need not be a the blank canvas either, instead we might wish to start from a photograph or an existing image, thus allowing continual evolution of images perhaps with different text conditioning on each iteration. The generative process may act on a 3D model which is then rendered to produce an image, rather than producing the image in 2D as we have done. The possibilities are endless for specifying a generative process as needed. In any case, the visual grammatical generative principle shown here can be extended to many domains of generation. \\

We should also be conscious of some biases in the multimodal transformer \citep{paullada2020data}\citep{mohamed2020decolonial}. For example, when we ask for `self-portrait' then the majority of portraits produced are of white males. Asking for a picture of a nurse or a doctor tends to replicate the biases present in pictures of these professions on the internet. Also, there is a more pervasive and less obvious bias in that rarer image classes such as ``black virgin and child'' are likely to be more challenging to score than common image classes, and so will produce worse images on average, and so may not be chosen by the designer to attempt. This is a more subtle effect of bias. \\

In conclusion, we believe this is an exciting new process which marks another shift in the relationship of the artist to their work. 
One of the authors (CF) who normally works in the medium of printmaking and oil paint describes his process with the system here.\\ 

\begin{quote}
``After developing the algorithm (and during its development) I spent weeks trying out various text prompts to see if it produced interesting images. I tried to understand what I was finding interesting in the images and modified the algorithm to maximize that probability. Initially I was trying to replicate human mark making, e.g.\ painterly images of clouds impressed me, apples that looked like apples, pears that looked like pears, etc... But then (after an informal ``crit'' with some friends at the Royal Collage of Art AI group) I accepted that I was judging the machine too much from my own biased perspective of what I thought good art was. I was not thinking about the potential of the machine itself, like a good art teacher might think about their student, or a parent about their child. I was also concerned about appropriation. For example, I had started to really enjoy the text prompt ``Tiger in the Jungle''. I was aware that much of the automated critic's influence must have been Rousseau. It was appropriating Rousseau. Then an artist friend told me Rousseau had never left France and that his wonderful painting in the Tate of ``Tiger in a Tropical Storm (Surprised)'' which I'd seen since I was a child and which is such an iconic image, which the multimodal transformer itself appropriates at a distanced and disembodied way from the internet, was kind of Rousseau's own distanced and disembodied view of tigers in jungles. I then remembered I'd actually been born in Sri-Lanka (a fact I often forget) and came to England at the age of 4. I'd seen a lot of jungle, not so many tigers. I had the urge to turn the phrase around and make it “Jungle in the Tiger”, and when I did that it moved me because it brought a sudden memory of feelings as a child emigrating from Sri-Lanka aged 4 because I grew up in a string of old Victorian psychiatric hospitals in the middle of marsh land (my father had been a psychiatrist) until I was about 9 when we moved to a normal house. They typically had big bath tubs in rooms with hugely high ceilings, and water that was dangerously hot. In the bath I had the recurring day dream of being in a tropical paradise, humid, dense foliage, with Bruce Springsteen playing a prehistoric song of not being able to start a fire without a spark. If I was the tiger, the jungle was still there in Wales in this lunatic asylum perhaps at least at bath times. That is what Jungle in the Tiger conveyed to me. Ok, so that was a whole load of associations, but I wondered what would happen if I put this variation of “Tiger in the Jungle” into the algorithm. As Figure \ref{fig:0} shows it came up with a 1950s Pulp Cinema Poster feeling. Not at all capturing any of the above, but which I loved. I have no idea what its done half the time, I think its tried to make a Tiger from a Jungle, and a Jungle from a Tiger. I couldn't have predicted the kind of thing it came up with. Anyway I was painfully aware that the multimodal transformer didn't know anything about my personal associations from which I might have created a piece of work about what ``Jungle in the Tiger'' means to me. Instead, it came up with something which is in a way its own interpretation of whatever that phrase means to itself, having had quite a different ``childhood'' reading the whole of the internet. Quite an unemotional childhood at that, with no engagement in the world as a reinforcement learning agent whatsoever that would allow it to link images through their emotions and not just through their associated words. God knows what it chose and why; that remains a mystery to me. And so when it produced the images they were a surprise. My role has been only to provide the generative process and the text prompt, and then to select, and to be disappointed most of the time. I've actually taken 16 consecutive independent evolutionary runs without censorship. I wanted to make the boundaries of my own critical choices clear at least. Maybe people will look at parts of it and see obvious cultural references. But I am pleased with the diversity of solutions that maximize “Jungle in the Tiger” as opposed to the similar solutions it finds to “Tiger in the Jungle”. Having said that, I’ve always loved Rousseau, and I think this piece will go together with a similar piece called “Tiger in the Jungle”, and they will work as a symmetric pair. I have to make further decisions about how to arrange the pieces, and so on, all not made by the algorithm, yet. This is not the final piece yet, which may change a lot. Coming back to this two days later, I no longer like what I've done at all and think it's a mess, and I seriously doubt my critical ability. A week later I know what I need to do to make it much better.''
\end{quote}

\section*{Acknowledgements} 
Thanks to Lizzie Mitchell, Eugenie Vronskaya, Claire Zakiewicz, Elizabeth\\ Hilliard Selka, Anja Borowicz Richardson, Rose Gibbs, Ian Woods, Andrew Mania, and Tom Stepleton for criticism and discussions. Thanks to Ben Coppin, Charlie Beattie, Jordan Hoffmann, Andrei Kashin, Thomas K\"{o}ppe, and all members of the generative art group for tremendous support.

\bibliographystyle{plain}
\bibliography{main}

\end{document}